\documentclass{article}
\usepackage{frExamplee}
\usepackage{graphicx}
\usepackage{apalike}
\usepackage{setspace}

\usepackage[ruled,vlined,linesnumbered]{algorithm2e}
\usepackage{hyperref}
\usepackage{siunitx}

\usepackage{amsmath}
\usepackage{gensymb}
\usepackage{amssymb}
\usepackage{makecell}

\usepackage{color}
\usepackage{soul}

\newcommand{\ie}{\textit{i}.\textit{e}.}
\newcommand{\eg}{\textit{e}.\textit{g}.}

\title{LiDAR-based Quadrotor Autonomous Inspection System in Cluttered Environments}

\author{
Wenyi Liu\thanks{These authors contributed equally.} \\
Department of Mechanical Engineering \\
University of Hong Kong \\
Pokfulam, Hong Kong \\
\texttt{liuwenyi@connect.hku.hk} \\
\And
Huajie Wu\footnotemark[1] \\
Department of Mechanical Engineering \\
University of Hong Kong \\
Pokfulam, Hong Kong \\
\texttt{wu2020@connect.hku.hk} \\
\AND
Liuyu Shi \\
Department of Mechanical Engineering \\
University of Hong Kong \\
Pokfulam, Hong Kong \\
\texttt{liuyushi@connect.hku.hk} \\
\And
Fangcheng Zhu \\
Department of Mechanical Engineering \\
University of Hong Kong \\
Pokfulam, Hong Kong \\
\texttt{zhufc@connect.hku.hk} \\
\And
Yuying Zou \\
Department of Mechanical Engineering \\
University of Hong Kong \\
Pokfulam, Hong Kong \\
\texttt{zyycici@connect.hku.hk} \\
\And
Fanze Kong \\
Department of Mechanical Engineering \\
University of Hong Kong \\
Pokfulam, Hong Kong \\
\texttt{kongfz@connect.hku.hk} \\
\And
Fu Zhang \\
Department of Mechanical Engineering \\
University of Hong Kong \\
Pokfulam, Hong Kong \\
\texttt{fuzhang@hku.hk} \\
}

\begin{document}

\maketitle

\begin{abstract}

In recent years, autonomous unmanned aerial vehicle (UAV) technology has seen rapid advancements, significantly improving operational efficiency and mitigating risks associated with manual tasks in domains such as industrial inspection, agricultural monitoring, and search-and-rescue missions. Despite these developments, existing UAV inspection systems encounter two critical challenges: limited reliability in complex, unstructured, and GNSS-denied environments, and a pronounced dependency on skilled operators. To overcome these limitations, this study presents a LiDAR-based UAV inspection system employing a dual-phase workflow: human-in-the-loop inspection and autonomous inspection. During the human-in-the-loop phase, untrained pilots are supported by autonomous obstacle avoidance, enabling them to generate 3D maps, specify inspection points, and schedule tasks. Inspection points are then optimized using the Traveling Salesman Problem (TSP) to create efficient task sequences. In the autonomous phase, the quadrotor autonomously executes the planned tasks, ensuring safe and efficient data acquisition. Comprehensive field experiments conducted in various environments, including slopes, landslides, agricultural fields, factories, and forests, confirm the system's reliability and flexibility. Results reveal significant enhancements in inspection efficiency, with autonomous operations reducing trajectory length by up to 40\% and flight time by 57\% compared to human-in-the-loop operations. These findings underscore the potential of the proposed system to enhance UAV-based inspections in safety-critical and resource-constrained scenarios.

\end{abstract}

\section{Introduction} 

Over the past decade, advancements in autonomous unmanned aerial vehicle (UAV) technology have led to their widespread adoption across various industries, including industrial inspections \cite{jordan2018state,bolourian2020lidar,du2022uav}, precision agriculture \cite{velusamy2021unmanned}, delivery services \cite{betti2024uav}, and search and rescue operations \cite{lyu2023unmanned}. Integrating UAVs into industrial applications has resulted in numerous benefits, such as reduced economic costs, enhanced operational efficiency, and mitigated risks to human workers.

These advantages are particularly pronounced in the field of inspections, where many tasks pose significant hazards to human workers, such as inspections in mining zones \cite{ren2019review} or in areas that are difficult to access, such as slopes \cite{liu2024lidar}. UAVs offer a safer and more efficient solution to these challenges, as they can easily access and navigate these areas. Furthermore, UAVs are capable of carrying various sensors, including optical cameras, Light Detection and Ranging (LiDAR) sensors, thermal imaging cameras, and others, to capture detailed information about inspection sites and transmit the data to users for analysis.

A typical UAV inspection system involves two primary stages: scheduling the inspection tasks and executing the designated tasks. Initially, the user assigns the areas of interest for inspection to the UAV. Subsequently, the UAV determines an optimal path to reach these locations and follows this path to monitor the specified areas. 

There are two common approaches for scheduling inspection tasks. The first method involves directly setting the inspection points. Users can designate points on a map obtained through pre-flight to scan the inspection sites \cite{elios3,dronut,mavic3,skydio2+,jung2019toward}, or by manually controlling the UAV to fly near the inspection points and record them \cite{santos2023unmanned}. This method relies heavily on experienced pilots, leading to high human costs. The second approach involves providing the features of the inspection points, allowing the UAV to explore the environment and searching for these features to locate the inspection points \cite{guan2021uav}. However, achieving accurate matches between the feature information and the actual sensor data can be challenging due to variations in illumination and environmental conditions.

To monitor the designated inspection area, several key modules are included in the inspection system, such as localization, mapping, planning, and control. Among these, the localization module is of paramount importance. Most commercial drones (\eg, DJI \cite{mavic3}, and Skydio \cite{skydio2+}) primarily rely on the Global Positioning System (GPS) for positioning. However, in GPS-denied environments, such as indoor settings, or areas with weak GPS signals, like dense forests, this reliance can pose challenges. To address this issue, some drones utilize vision-based localization \cite{omari2014visual,grando2020visual,nikolic2013uav}, enabling indoor operations but facing difficulties in low-light conditions and with detecting thin obstacles, such as branches, which can compromise flight safety.

\begin{figure}[h]
    \centering
    \includegraphics[width=\textwidth]{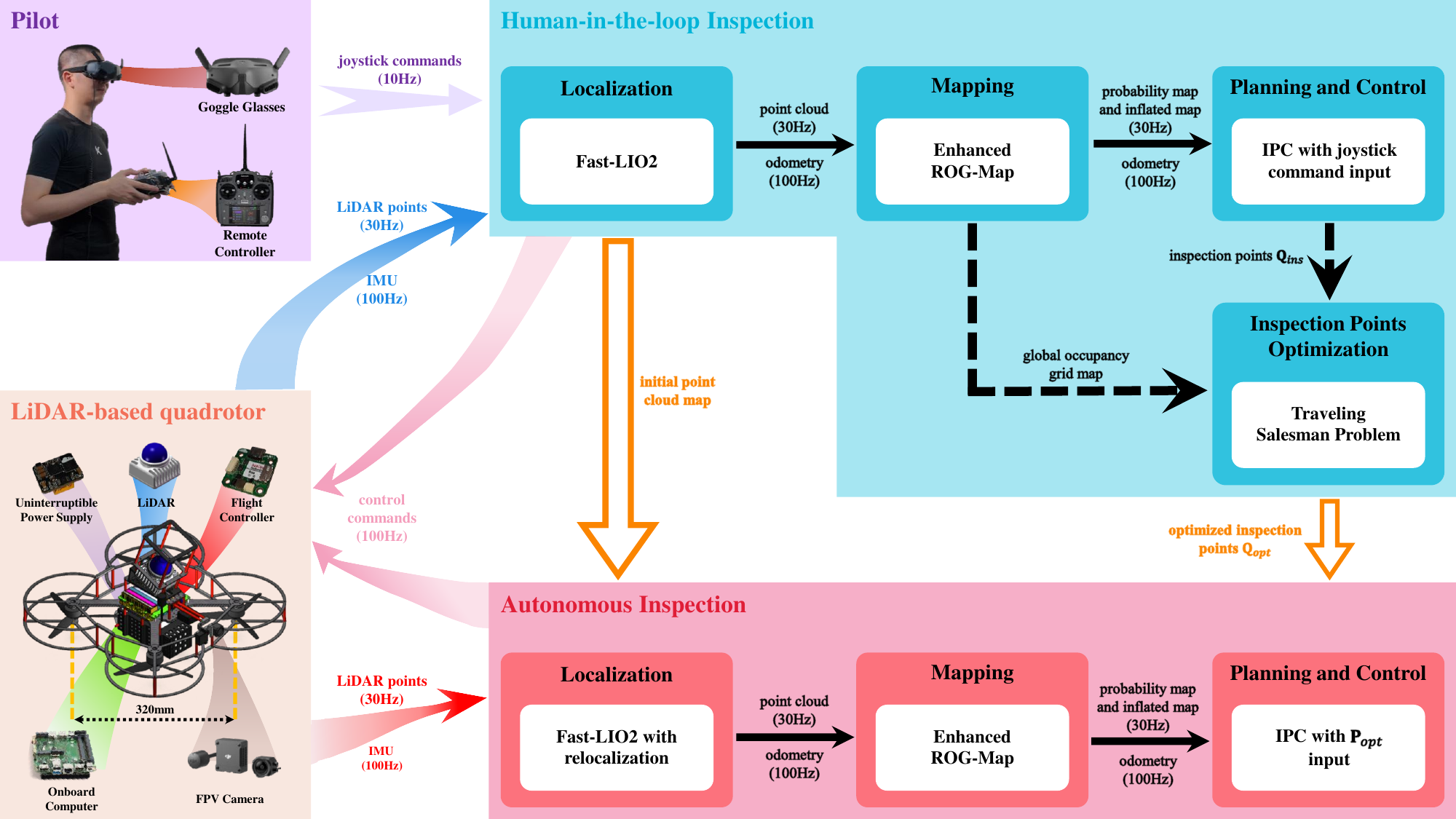}
    \caption{The system overview of our LiDAR-based quadrotor.}
    \label{fig:system}
\end{figure}

To address the aforementioned challenges, this article presents a comprehensive LiDAR-based inspection system, consisting of two main components: human-in-the-loop inspection and autonomous inspection, as illustrated in Fig. \ref{fig:system}. In the human-in-the-loop inspection phase, users can scan the environment to get the 3D map and schedule inspection tasks. Subsequently, in the autonomous inspection phase, the UAV executes these tasks independently. The system offers two significant advantages: it can be operated by untrained pilots and it can perform various inspection tasks autonomously even in unknown, GNSS-denied, and cluttered environments. It is worth noting that this work focuses on implementing and validating the integrated and complete system. As for the technical details of individual modules, prior work (\eg, FAST-LIO2 \cite{xu2022fast} in Localization module, ROG-Map \cite{ren2023rog} in Mapping module, IPC \cite{liu2023integrated} in Planning and Control module) is adopted here.

During the human-in-the-loop inspection phase, the system is used to scan a 3D map of the unknown environment, which can be used to schedule inspection tasks. An autonomous obstacle avoidance mechanism assists the pilot in controlling the quadrotor during this phase, significantly lowering the threshold for system use. Even an untrained pilot can easily operate the system. Specifically, the pilot simply needs to direct the quadrotor to target points and record the position, yaw angle, and gimbal pitch angle at each point, without worrying about collision between the UAV and the environment, as the UAV autonomously avoids obstacles in flight. After the flight, all recorded inspection points are used to optimize and find the shortest path that can traverse these points by solving the Traveling Salesman Problem (TSP).

The autonomous inspection phase is designed to execute the scheduled inspection tasks. During this stage, the quadrotor autonomously follows the optimized path derived from TSP to reach each inspection point in sequence, recording sensor data such as images and LiDAR information for detailed analysis. Notably, we have employed LiDAR on the UAV and integrated Simultaneous Localization and Mapping (SLAM) technology into the localization module. This LiDAR-based navigation design enables reliable operation in unknown, cluttered, and GNSS-denied environments and enhances overall inspection efficiency and safety.

We validated the proposed system by performing various tasks in cluttered, unstructured, and GNSS-denied environments, including the slope, landslide, agricultural area, factory, and forest. All experiments were conducted in real-world settings. Initially, the human-in-the-loop inspection phase was performed to obtain inspection points, followed by the autonomous phase to monitor these points sequentially. Flight trajectories from both phases were illustrated and compared based on metrics such as maximum speed, trajectory length, and flight time. The autonomous inspection phase achieved up to 40\% shorter trajectory lengths and 57\% reduced flight time, with a comparable maximum speed, compared to the human-in-the-loop inspection phase. These results demonstrate that our inspection system can monitor inspection points autonomously, with lower energy consumption and better efficiency.

\section{Hardware Structure}
\label{sec:hardware}

\begin{figure}[h]
    \centering
    \includegraphics[width=\textwidth]{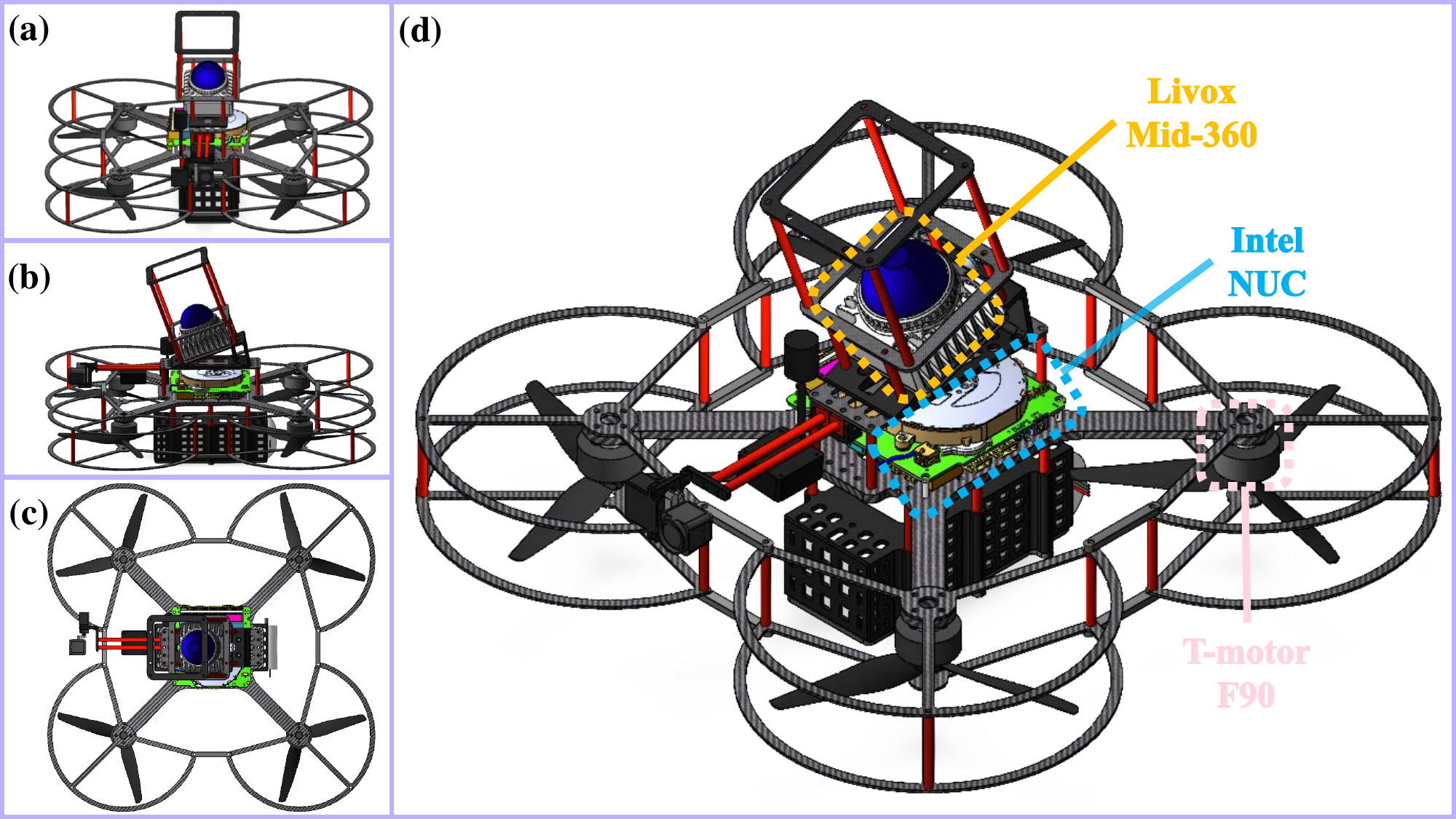}
    \caption{Different views of our LiDAR-based quadrotor.}
    \label{fig:uav}
\end{figure}

Table \ref{tab:hardware} and Fig. \ref{fig:uav} provide a comprehensive description of the hardware configuration of our LiDAR-based quadrotor. The quadrotor features a lightweight Livox Mid-360 LiDAR\footnote{\href{https://www.livoxtech.com/mid-360}{https://www.livoxtech.com/mid-360}}, weighing only \SI{265}{g}, as its primary sensor. This LiDAR employs a non-repetitive scanning approach, accumulating data over time to generate dense point cloud maps. Besides, it features a 360-degree horizontal field of view and a 59-degree vertical field of view, enabling the perception of a wide range of scenes. The onboard computation is handled by an Intel NUC mini-computer, featuring an Intel i7-1260P CPU capable of operating at a frequency of \SI{4.7}{GHz}. This high-performance processor provides sufficient computational power for real-time processing tasks. To enable real-time observation of the flight process by the pilot and capture photos and videos of specific areas, we use an FPV camera DJI O3 Air Unit\footnote{\href{https://www.dji.com/o3-air-unit}{https://www.dji.com/o3-air-unit}}, a high-definition digital video transmission system, with the DJI Goggles 2\footnote{\href{https://www.dji.com/goggles-2}{https://www.dji.com/goggles-2}}. This video transmission system offers an impressively low latency of \SI{40}{ms}, providing timely feedback on the surrounding environment for the pilot. Moreover, it supports recording \SI{4}{K} videos at a high frame rate of \SI{120}{Hz}, enabling in-depth post-analysis after the flights. To enlarge the Field of View of the FPV camera and facilitate monitoring both the top and bottom of the flexible debris-resisting barriers on the slope, the camera is installed on a pitch-axis gimbal. The pitch angle of the gimbal is commanded by the remote controller in real-time during flight. To maximize flight endurance, which is a critical consideration, our quadrotor is equipped with 7-inch propellers and a high-capacity 6S-5300mAh battery. To enhance quadrotor safety, carbon fiber propeller guards are meticulously designed and installed to minimize the risk of propeller-related accidents. These guards protect both the quadrotor and the pilot, ensuring safe and reliable operation. Furthermore, we design a flight controller based on the PX4 FMUv4\footnote{\href{https://docs.px4.io/main/en/flight_controller/}{https://docs.px4.io/main/en/flight\_controller/}} standard, which reduced both the cost and size of our quadrotr system. To reduce deployment time, we develop an uninterruptible power supply (UPS) module, ensuring uninterrupted power supply to onboard computers and electronics during battery replacement.

\begin{table}[h]
    \centering
    \caption{Device Information of our In-House Developed Quadrotor.}
    \label{tab:hardware}
    \begin{tabular}{|c|c|c|}
        \hline
        Device & Description & Weight (g) \\
        \hline
        ESC & T-motor F60A 8S 4IN1 & 15.3 \\
        \hline
        Motor & T-motor F90 KV1300 & 41.8 \\
        \hline
        Propeller & DALPROP T7057 & 7.5 \\
        \hline
        Receiver & RadioLink R12DSM & 2.5 \\
        \hline
        Flight Controller & In-house designed & 6.3 \\
        \hline
        Uninterruptible Power Supply & In-house designed & 2.7 \\
        \hline
        Battery & ACE 6S-5300mAh-30C lithium battery & 664 \\
        \hline
        Onboard Computer & Intel NUC with Intel i7-1260P CPU & 270 \\
        \hline
        LiDAR & Livox Mid-360 & 265 \\
        \hline
        FPV Camera & DJI O3 Air Unit & 36.4 \\
        \hline
        Goggle glasses & DJI Goggles 2 & 290 \\
        \hline
        Remote Controller & RadioLink AT10S PRO & 980 \\
        \hline
    \end{tabular}
\end{table}

After these devices are integrated into the airframe composed of carbon plates and aluminum columns, we conduct tests on the developed quadrotor. The quadrotor features a motor-to-motor distance (\ie, wheelbase) of \SI{320}{mm} and overall dimensions of \SI{422}{mm} in length, \SI{422}{mm} in width, and \SI{270}{mm} in height. With a total mass of \SI{2}{kg}, the quadrotor achieves a thrust-to-weight ratio of 3, ensuring efficient and stable flight performance. Additionally, {the maximum flight endurance} is measured to be 12 minutes.

\section{Software Structure}
\label{sec:software}

The software structure of our LiDAR-based quadrotor is shown in Fig. \ref{fig:system}. The software architecture comprises two primary components: human-in-the-loop inspection (Sec. \ref{sec:human}), and autonomous inspection (Sec. \ref{sec:auto}). 

In the human-in-the-loop inspection phase, the quadrotor is guided by the pilot's joystick commands and data inputs from onboard sensors (\eg, LiDAR and IMU) to perform inspection tasks in an unknown and cluttered environment. The pilot controls the quadrotor to reach the inspection points and records the position {$\mathbf{p}$}, yaw angle {$\phi$}, and gimbal pitch angle {$\theta$} at each inspection point {$\mathbf{q}=(\mathbf{p},\phi,\theta)$}. Simultaneously, the quadrotor autonomously avoids obstacles during flight to ensure flight safety. After completing the human-in-the-loop inspection flight, inspection points optimization (Sec. \ref{sec:tsp}) solves the Traveling Salesman Problem (TSP) to reorder the recorded {$N$} inspection points {$\mathbf{Q}_{ins}=[\mathbf{q}_{ins,0},\mathbf{q}_{ins,1}, ... ,\mathbf{q}_{ins,N-1}]$} based on the shortest path criterion. The optimized {$N$} inspection points {$\mathbf{Q}_{opt}=[\mathbf{q}_{opt,0},\mathbf{q}_{opt,1}, ... ,\mathbf{q}_{opt,N-1}]$}, are subsequently used as inputs for the autonomous inspection phase.

In the autonomous inspection phase, the quadrotor, relying on LiDAR and IMU sensors, autonomously executes repetitive inspection tasks within the environment previously explored during the human-in-the-loop inspection phase. The quadrotor navigates to the optimized inspection points in sequence, collecting image and LiDAR data for subsequent detailed analysis.

\subsection{Human-in-the-loop Inspection}
\label{sec:human}

\subsubsection{Localization}
\label{sec:human_lio}

In human-in-the-loop inspection, the localization module sets the initial position $\mathbf{p}_{init}$ of the quadrotor as the origin of the world frame to obtain the corresponding LiDAR-inertial odometry. In this work, the open-source framework FAST-LIO2 \cite{xu2022fast} is adapted to meet the system requirements. By performing IMU pre-integration, the modified FAST-LIO2 provides low-latency state estimation at a frequency of \SI{100}{Hz}, with latency under \SI{1}{ms}. Moreover, it generates world-coordinate-registered point clouds at a scanning rate of \SI{30}{Hz}, producing approximately 200,000 points per second with a latency of less than \SI{10}{ms}.

To further support the relocalization function during subsequent autonomous inspection tasks, an initial point cloud map is generated for feature matching. The quadrotor remains stationary at its initial position $\mathbf{p}_{init}$ while accumulating LiDAR point cloud data over a period (\eg, \SI{5}{s}). Leveraging the non-repetitive scanning characteristic of the Mid-360 LiDAR, this process generates a dense and feature-rich initial point cloud map.



\subsubsection{Mapping}
\label{sec:human_map}

The open-source framework ROG-Map \cite{ren2023rog} utilizes a zero-copy map sliding strategy to maintain two local occupancy grid maps (OGMs). The first local map is the probability map, where grid cells are classified as Occupied, Unknown, or Known Free based on the occupancy probability updates derived from ray casting. The second local map is the inflated map, utilized for robot navigation in configuration space by inflating obstacles. This map adopts an incremental update mechanism, with grid cell states classified as either Inflation or No Inflation, depending on the probability values from the probability map.

In \cite{liu2024lidar}, three enhancements were introduced to ROG-Map specifically tailored for dense vegetation environments: Unknown Grid Cells Inflation, Infinite Points Ray Casting, and Incremental Frontiers Update. Unknown Grid Cells Inflation expands Unknown grid cells to the robot's size, preventing collisions with obstacles in Unknown areas and significantly improving safety. Infinite Points Ray Casting tackles the issue of no LiDAR returned points when facing the sky. Incremental Frontiers Update efficiently updates frontier information based on the latest sensor data. This process replaces a large number of Unknown grids with a small number of frontier grids, thereby reducing the computation time required for safe flight corridor (SFC) generation in the planning. These three enhancements are applicable across various inspection tasks, so we directly adopted this enhanced version of ROG-Map. For more technical details, please refer to \cite{liu2024lidar}.

In addition, we maintain an additional large-scale, low-resolution global occupancy grid map. This map is centered on the initial position $\mathbf{p}_{init}$ and remains static, without sliding relative to the robot's position. Each grid cell in this map represents either an Occupied or Free state, determined based on the grid state in the probabilistic map. For example, if a grid cell in the probabilistic map is marked as Known Free, its state is set to Free; otherwise, it is considered Occupied. This global map serves as the foundation for path planning in subsequent inspection points optimization, specifically for determining the path lengths between pairs of inspection points.

\subsubsection{Planning and Control}
\label{sec:human_ipc}

In human-in-the-loop inspection, the pilot determines the inspection targets impromptu from the transmitted video and commands the quadrotor's high-level flight direction via the joysticks on the remote controller. Meanwhile, the quadrotor navigation system attempts to fly in the specified direction while avoiding obstacles on the way. During the flight, the pilot can use the joystick on the remote controller to record the current position, orientation and camera angle, which will serve as the inspection point $\mathbf{q}=(\mathbf{p},\phi,\theta)$.

We directly adopt {the integrated planning and control (IPC) module} from \cite{liu2024lidar} to achieve obstacle avoidance functionality for our quadrotor. IPC first calculates the local goal and yaw reference of the quadrotor based on the pilot's joystick commands and odometry, then searches for the reference path on the inflated map. Following this, the safety flight corridor (SFC) is generate on the reference path within the Known Free area of the probability map. The reference path and SFC are then used as positional references and safety constraints, respectively, for the Model Predictive Control (MPC) problem. Subsequently, the optimal control actions of the MPC are transformed into the quadrotor actuator commands through differential flatness \cite{mellinger2011minimum}, which are finally tracked by a lower-level controller implemented on the autopilot to produce motor commands.


\subsubsection{Inspection Points Optimization}
\label{sec:tsp}

\begin{figure}[h]
    \centering
    \includegraphics[width=\textwidth]{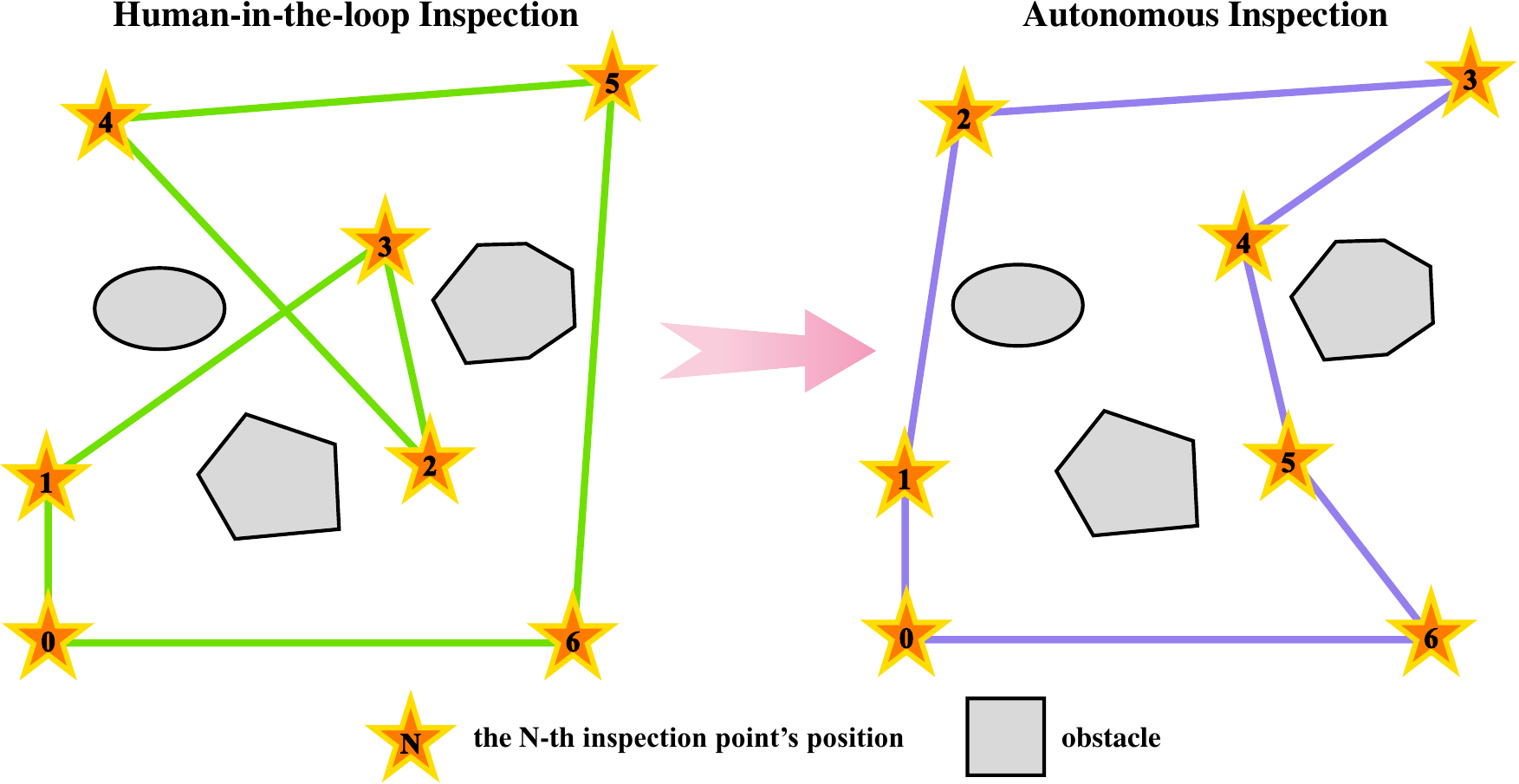}
    \caption{Inspection points optimization in the 2D case. By solving the Traveling Salesman Problem (TSP), the sequence of inspection points under human-in-the-loop inspection (green lines with yellow stars) is optimized, resulting in the shortest total path (purple lines with yellow stars), thereby significantly improving efficiency in autonomous inspection.}
    \label{fig:tsp}
\end{figure}

During the human-in-the-loop inspection flight, $N$ inspection points $\mathbf{Q}_{ins}=[\mathbf{q}_{ins,0},\mathbf{q}_{ins,1}, ... ,\mathbf{q}_{ins,N-1}]$ are recorded, but their spatial distribution in the three-dimensional space may be scattered, leading to intersecting paths that significantly reduce inspection efficiency in autonomous inspection, as illustrated by the green lines with yellow stars in Fig. \ref{fig:tsp}. To enhance inspection efficiency, it is necessary to find the shortest path that visits each inspection point once and returns to the starting point (depicted by the purple lines with yellow stars in Fig. \ref{fig:tsp}), given a series of inspection points and the distances between each pair of points. This problem is equivalent to the well-known Traveling Salesman Problem (TSP) \cite{junger1995traveling,meng2017two}, for which numerous solution methods \cite{shi2007particle,deudon2018learning} exist. Inspired by \cite{zhou2021fuel,zhou2023racer}, we employ a Lin-Kernighan-Helsgaun heuristic solver \cite{helsgaun2000effective} to solve the TSP in real time. Additionally, since the Euclidean distance between two inspection points in cluttered environments may differ significantly from the actual navigation path length, we utilize the A* algorithm \cite{hart1968formal} on a pre-constructed global map to calculate the path length between any two inspection points. Finally, we obtain the optimized sequence of inspection points {$\mathbf{Q}_{opt}=[\mathbf{q}_{opt,0},\mathbf{q}_{opt,1}, ... ,\mathbf{q}_{opt,N-1}]$}, which serves as targets for the quadrotor in autonomous inspection.

\subsection{Autonomous Inspection}
\label{sec:auto}

\subsubsection{Localization}
\label{sec:auto_lio}

In autonomous inspection, the localization module continues to utilize the modified Fast-LIO2 (Sec. \ref{sec:human_lio}) to estimate the full state of the quadrotor and transform LiDAR point clouds from the local frame to the world frame. To maintain consistency in the inspection points between human-in-the-loop inspection and autonomous inspection, the localization module in this mode incorporates an additional relocalization function to align the odometric world frame within the same inspection site. This alignment is achieved using the point-to-plane Iterative Closest Point (ICP) algorithm \cite{besl1992method}, which provides an initial state estimate by relocalizing the quadrotor near the origin of the initial point cloud map (\ie, initial position $\mathbf{p}_{init}$) constructed during human-in-the-loop inspection (Sec. \ref{sec:human_lio}).

Given the sensitivity of ICP to rotation, directly performing full six-degrees-of-freedom (6-DOF) registration initially may not converge. Instead, we sample a range of initial rotations, only optimizing the translation for each. The rotation and optimized translation corresponding to the lowest cost are recorded as the initial pose, followed by a full 6-DOF ICP to determine the quadrotor's {current position} in the initial point cloud map.


\subsubsection{Mapping}
\label{sec:auto_map}

The mapping module in autonomous inspection, similar to the human-in-the-loop inspection (Sec. \ref{sec:human_map}), utilizes the method from \cite{liu2024lidar} to maintain the probability map and inflated map. However, a key difference lies in the fact that, since the inspection points have already been optimized, the mapping module in this mode does not need to maintain an additional global occupancy grid map.


\subsubsection{Planning and Control}
\label{sec:auto_ipc}

In autonomous inspection, we adopt the work \cite{liu2023integrated} to enable autonomous navigation for the quadrotor. Starting from its current position, the quadrotor sequentially takes each inspection point's position in $\mathbf{q}_{opt,n}=[\mathbf{p}_{opt,n},\phi_{opt,n},\theta_{opt,n}]$ for $n = 0, 1, ..., N - 1$ as the target and uses the A* algorithm \cite{hart1968formal} to search for reference paths. Following the SFC generation method from \cite{liu2017planning}, two or more corridors are created along the reference path to serve as the SFC. Upon reaching each destination, the quadrotor hovers for a preset duration (\eg, 3 seconds) and adjusts its yaw angle and gimbal pitch angle according to the record $\Psi_{opt}$ and $\Theta_{opt}$, respectively, to capture images of the specified area.

However, in some scenarios, environmental changes such as falling branches or moving objects may render the inspection points unsafe (\ie, in Unknown Inflation or Occupied Inflation region of the inflated map). In such cases, the quadrotor performs a breadth-first search to locate the nearest No Inflation point as a hover point. Although the quadrotor may not reach the exact predefined inspection target, it still adjusts its yaw and gimbal pitch angles to collect image data. If the quadrotor fails to reach the predefined inspection point and the hover time exceeds a preset limit (\eg, 5 seconds), the inspection point is abandoned, and a new reference path is generated from the current position to the next inspection point. This approach ensures that the quadrotor can autonomously complete the entire inspection task and return to the first point in $\mathbf{P}_{opt}$ to land.


\section{Field Experiments}
\label{sec:experiments}

To validate the practicality of our LiDAR-based quadrotor inspection system for various tasks, we conducted field experiments in multiple cluttered and unstructured environments, including scenarios such as slope, landslide, agriculture, factory, and forestry. These tasks all follow the workflow of human-in-the-loop inspection followed by autonomous inspection, aiming to demonstrate the broad applicability of our approach beyond the built environment, particularly in scenarios where environmental conditions are uncontrollable.

In the first phase, the human-in-the-loop inspection, the quadrotor is deployed to a new inspection scene, where the pilot guides it using joystick commands. The quadrotor autonomously avoids obstacles along its flight path and reaches designated inspection targets. During this phase, the position, yaw angle, and gimbal pitch angle of the inspection points are recorded. Once the human-in-the-loop inspection is completed, the Traveling Salesman Problem (TSP) is solved to optimize the sequence of the recorded inspection points, including their positions and corresponding yaw and gimbal pitch angles, to generate the shortest possible inspection path.

In the second phase, the autonomous inspection, the quadrotor operates independently within the same environment. Using the optimized inspection point data, it autonomously collects images and LiDAR point cloud data from the inspection targets without requiring human intervention.

\subsection{Inspection for Slope Task}

\begin{figure}[h]
    \centering
    \includegraphics[width=\textwidth]{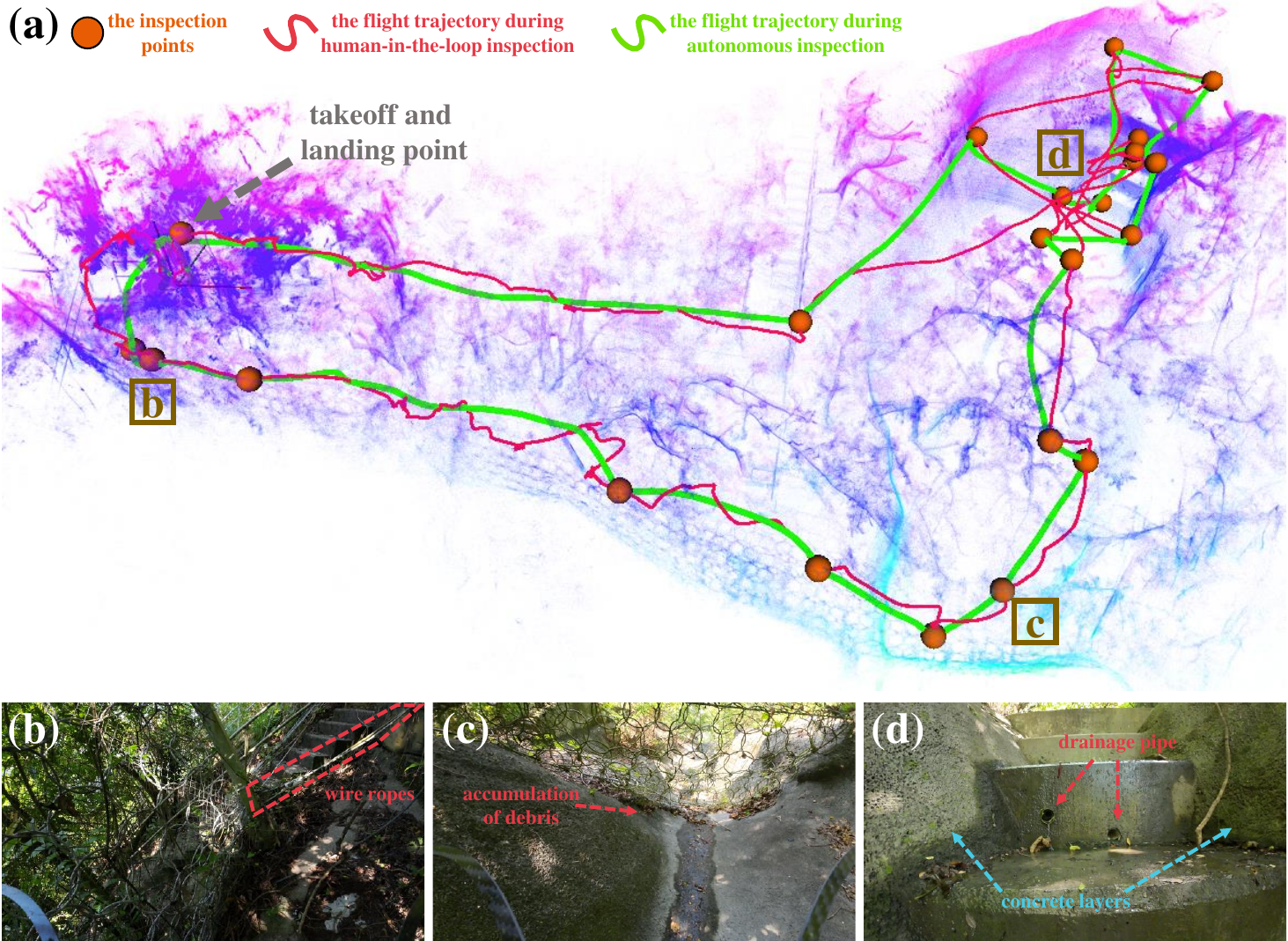}
    \caption{Overview of the quadrotor inspection system in slope task. (a) Point cloud map constructed during the inspection. (b) Image captured of the corrosion of the wire ropes supporting the barriers. (c) Image captured of the accumulation of debris behind barriers. (d) Image captured of the condition of the blocked drainage pipe and corrosion within the concrete layers.}
    \label{fig:slope}
\end{figure}

We conducted a quadrotor inspection task in Hong Kong, an advanced region for slope management\footnote{\href{https://www.bbc.com/future/article/20220225-how-hong-kong-protects-people-from-its-deadly-landslides}{https://www.bbc.com/future/article/20220225-how-hong-kong-protects-people-from-its-deadly-landslides}}. The specific site selected was the 11SW-C/ND6 slope near Victoria Road, Pokfulam, characterized by dense vegetation and steep terrain. The primary inspection targets included man-made structures, such as the concrete layers reinforcing the mountainside and flexible debris-resisting barriers designed to intercept landslide debris. The inspection focused on identifying the following issues: corrosion of the wire ropes supporting the barriers (Fig. \ref{fig:slope}(b)), accumulation of debris at the bottom of the barriers (Fig. \ref{fig:slope}(c)), and the condition of the blocked drainage pipe and corrosion within the concrete layers (Fig. \ref{fig:slope}(d)).

\begin{table}[h]
    \centering
    \caption{Flight Data of Inspection for Slope Task}
    \label{tab:slope_data}
    \begin{tabular}{|c|c|c|c|c|}
        \hline
        Inspection Mode & Maximum & Average & Trajectory & Flight Time \\
         & Speed $(m/s)$ & Speed $(m/s)$ & Length $(m)$ & $(min : sec)$ \\
        \hline
        Human-in-the-loop & 2.02 & 0.57 & 176.88 & 5 : 09 \\
        \hline
        Autonomous & 2.20 & 0.88 & 118.42 & 2 : 14 \\
        \hline
    \end{tabular}
\end{table}

In this experiment, our quadrotor completed the full inspection workflow. Flight data recorded during the inspection, as shown in Table \ref{tab:slope_data}, demonstrate the system’s efficiency in navigating through complex environments, with key metrics such as coverage volume, maximum flight speed, trajectory length, and flight time. Compared to the human-in-the-loop inspection phase, the autonomous inspection phase achieved a 33\% reduction in trajectory length and a 57\% reduction in flight time. This improvement is attributed to the cluttered and dense slope environment, where, during the human-in-the-loop inspection, the quadrotor frequently rejected coarse commands from inexperienced operators. In contrast, the autonomous inspection allowed the quadrotor greater freedom, enabling more precise and efficient navigation. 

\begin{figure}[h]
    \centering
    \includegraphics[width=\textwidth]{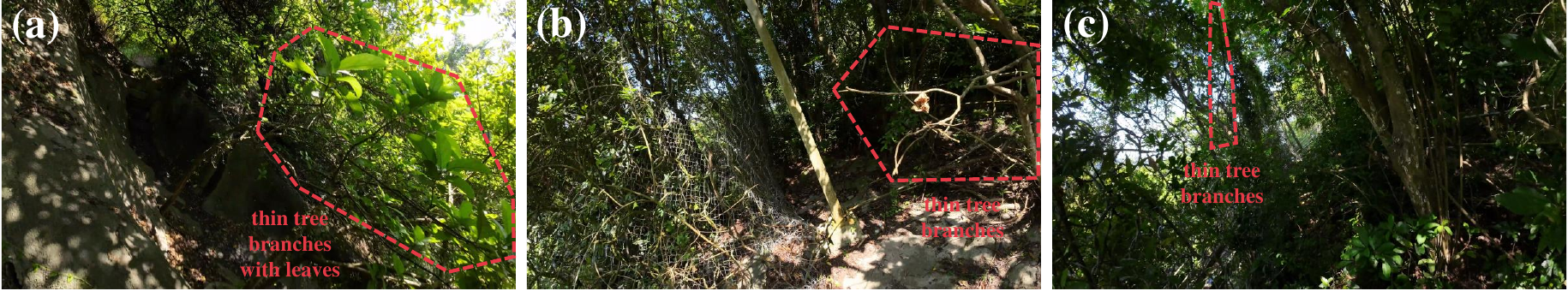}
    \caption{First-person view photos collected during autonomous inspection. Our quadrotor successfully avoided small obstacles, demonstrating the robustness of the navigation system.}
    \label{fig:slope_small}
\end{figure}

\begin{figure}[h]
    \centering
    \includegraphics[width=\textwidth]{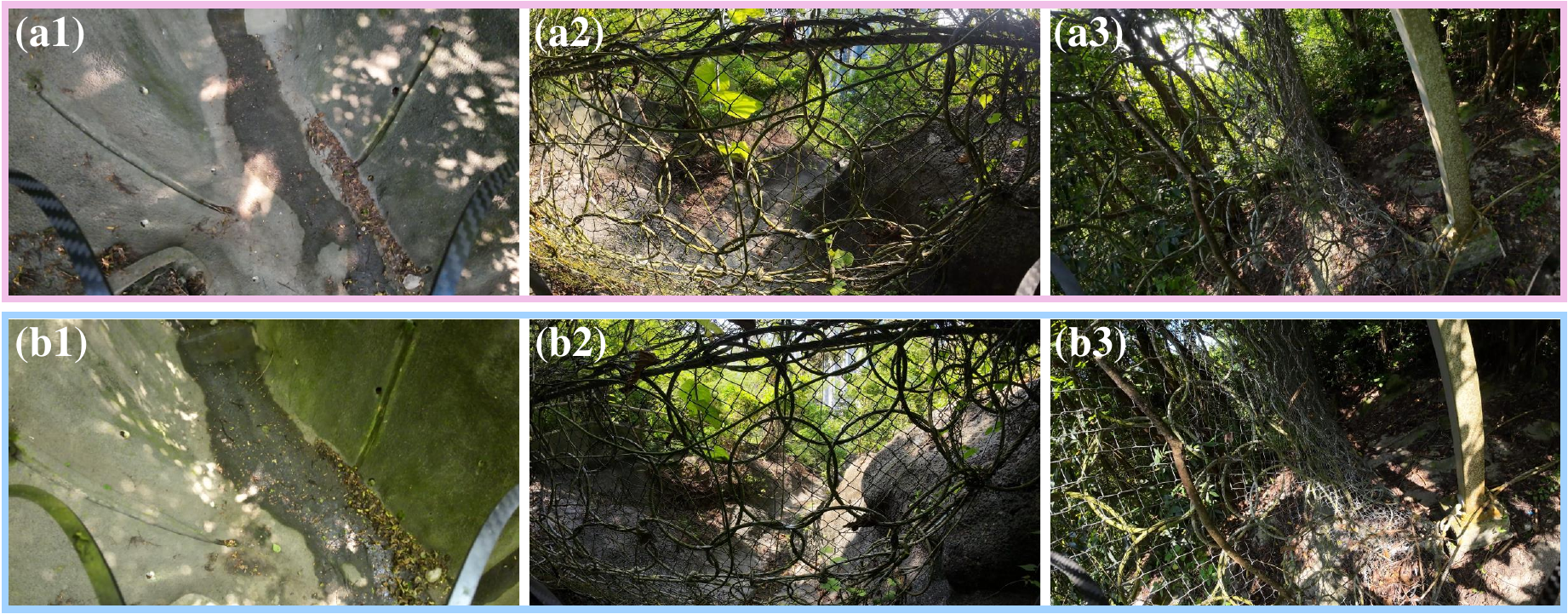}
    \caption{Comparison of photos collected at the same inspection points during human-in-the-loop inspection and autonomous inspection. (a) Images collected during human-in-the-loop inspection. (b) Images collected during autonomous inspection.}
    \label{fig:slope_bench}
\end{figure}

As shown in Fig. \ref{fig:slope}, during the autonomous inspection, the quadrotor autonomously navigated to inspection points recorded during the human-in-the-loop phase, following the shortest path to collect data. Additionally, it successfully avoided small obstacles (\eg, thin tree branches) in the environment (see Fig. \ref{fig:slope_small}), demonstrating the robustness of its onboard real-time navigation system. Furthermore, by leveraging the re-localization functionality of the localization module (Sec. \ref{sec:auto_lio}), the inspection target photos collected during both the human-in-the-loop and autonomous phases are nearly identical, as illustrated in Fig. \ref{fig:slope_bench}. Notably, when certain inspection points were obstructed by fallen branches or leaves, the quadrotor employed a breadth-first search algorithm to identify the nearest Known Free point for hovering (Sec. \ref{sec:auto_ipc}).

Through efficient quadrotor inspection, the latest conditions of the slope area were captured, potential safety hazards were promptly identified, and valuable data was provided to support slope management and maintenance efforts.

We invite the readers to watch our first supplementary video\footnote{\href{https://youtu.be/irb_lL9NQJU}{https://youtu.be/irb\_lL9NQJU}}, to gain a clearer view of the efficiency and flexibility of the quadrotor when performing slope inspection tasks.


\subsection{Inspection for Landslide Task}

\begin{figure}[h]
    \centering
    \includegraphics[width=\textwidth]{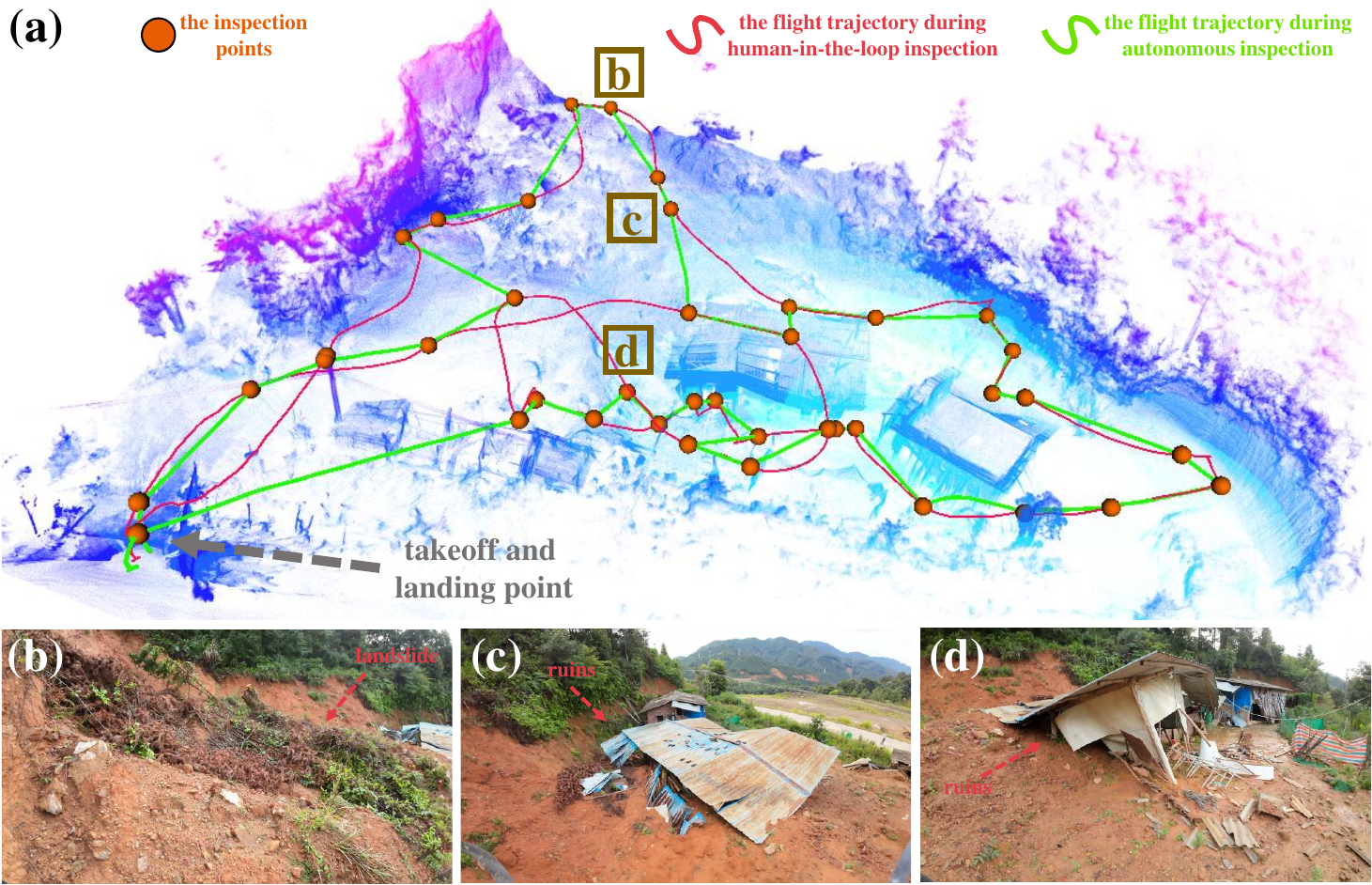}
    \caption{Overview of the quadrotor inspection system in landslide task. (a) Point cloud map constructed during the inspection. (b) Image captured of the surface characteristics of the landslide. (c-d) Image captured of the condition of the ruins.}
    \label{fig:landslide}
\end{figure}

On June 16, 2024, Pingyuan County in Guangdong Province, China, characterized by hilly terrain, experienced an extreme rainfall event\footnote{\href{https://news.cctv.com/2024/06/21/ARTIKUnHB6cYSIS0idoFnrVb240621.shtml}{https://news.cctv.com/2024/06/21/ARTIKUnHB6cYSIS0idoFnrVb240621.shtml}}. This severe weather led to natural disasters such as landslides and floods, causing significant casualties and property damage. We conducted a quadrotor inspection task in Sishui Township, the area most severely impacted by the disaster. In the landslide task, the primary inspection targets included the surface characteristics of the landslide (Fig. \ref{fig:landslide}(b)) and the condition of the ruins (Fig. \ref{fig:landslide}(c) and Fig. \ref{fig:landslide}(d)).

\begin{table}[h]
    \centering
    \caption{Flight Data of Inspection for Landslide Task}
    \label{tab:landslide_data}
    \begin{tabular}{|c|c|c|c|c|}
        \hline
        Inspection Mode & Maximum & Average & Trajectory & Flight Time \\
         & Speed $(m/s)$ & Speed $(m/s)$ & Length $(m)$ & $(min : sec)$ \\
        \hline
        Human-in-the-loop & 2.43 & 1.00 & 276.75 & 4 : 36 \\
        \hline
        Autonomous & 2.21 & 0.96 & 238.97 & 4 : 08 \\
        \hline
    \end{tabular}
\end{table}

In this experiment, our quadrotor completed the entire inspection workflow. The flight data recorded during the inspection is summarized in Table \ref{tab:landslide_data}. Due to the relatively open environment caused by the landslide, the maximum flight speed during the human-in-the-loop inspection phase was slightly higher than that of the autonomous inspection phase. Consequently, the autonomous inspection achieved only a 14\% reduction in trajectory length and a 10\% reduction in flight time.

\begin{figure}[h]
    \centering
    \includegraphics[width=\textwidth]{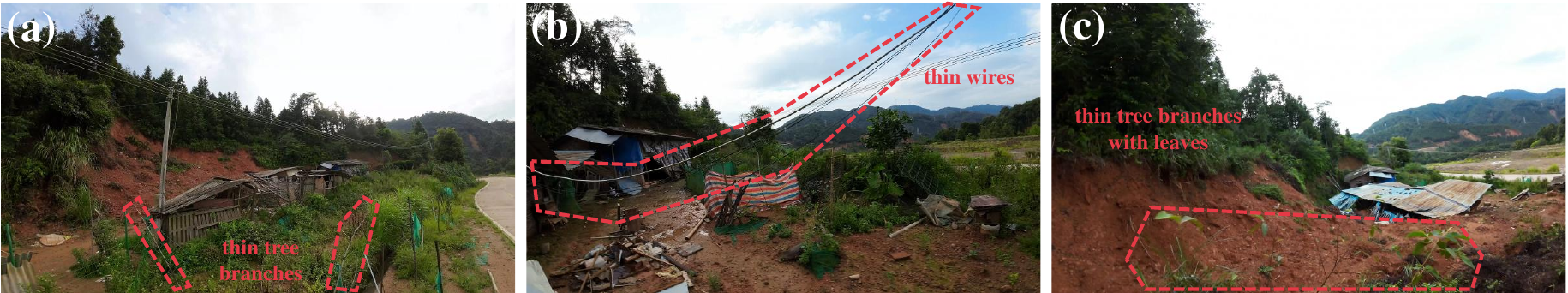}
    \caption{First-person view photos collected during autonomous inspection. Our quadrotor successfully avoided small obstacles, demonstrating the robustness of the navigation system.}
    \label{fig:landslide_small}
\end{figure}

\begin{figure}[h]
    \centering
    \includegraphics[width=\textwidth]{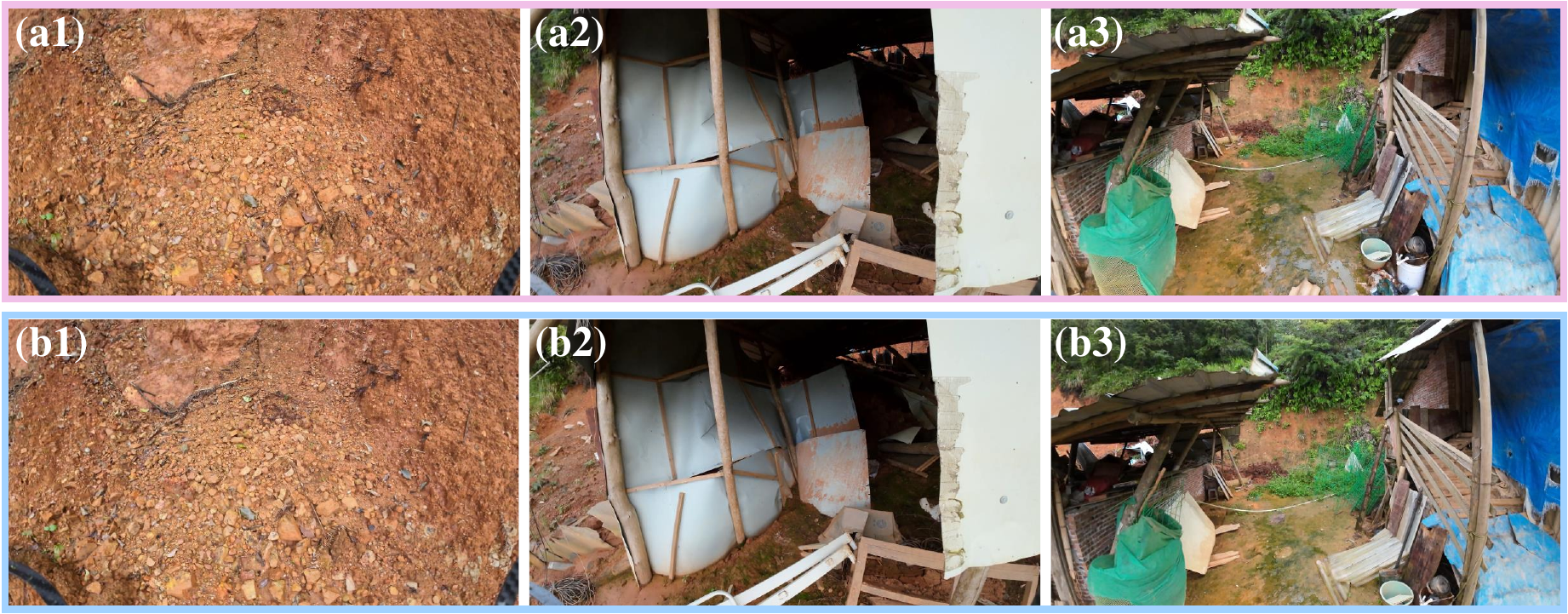}
    \caption{Comparison of photos collected at the same inspection points during human-in-the-loop inspection and autonomous inspection. (a) Images collected during human-in-the-loop inspection. (b) Images collected during autonomous inspection.}
    \label{fig:landslide_bench}
\end{figure}

As illustrated in Fig. \ref{fig:landslide}, during the autonomous inspection, the quadrotor autonomously navigated to the inspection points recorded during the human-in-the-loop phase, following the shortest path to collect data. Additionally, it safely avoided small obstacles (\eg, thin tree branches and wires) present in the environment (see Fig. \ref{fig:landslide_small}). By leveraging the re-localization functionality of the localization module (Sec. \ref{sec:auto_lio}), the photos of the inspection targets collected during both the human-in-the-loop and autonomous phases were nearly identical, as shown in Fig. \ref{fig:landslide_bench}.

Through efficient quadrotor inspection, we were able to accurately assess the latest conditions of the landslide-affected area, promptly identify potential safety hazards, and provide critical data support for subsequent rescue and restoration efforts.

We invite the readers to view our second supplementary video\footnote{\href{https://youtu.be/tDQllM3uo18}{https://youtu.be/tDQllM3uo18}} for a clearer understanding of the entire process of the quadrotor executing the landslide inspection task.


\subsection{Inspection for Agriculture Task}

\begin{figure}[h]
    \centering
    \includegraphics[width=\textwidth]{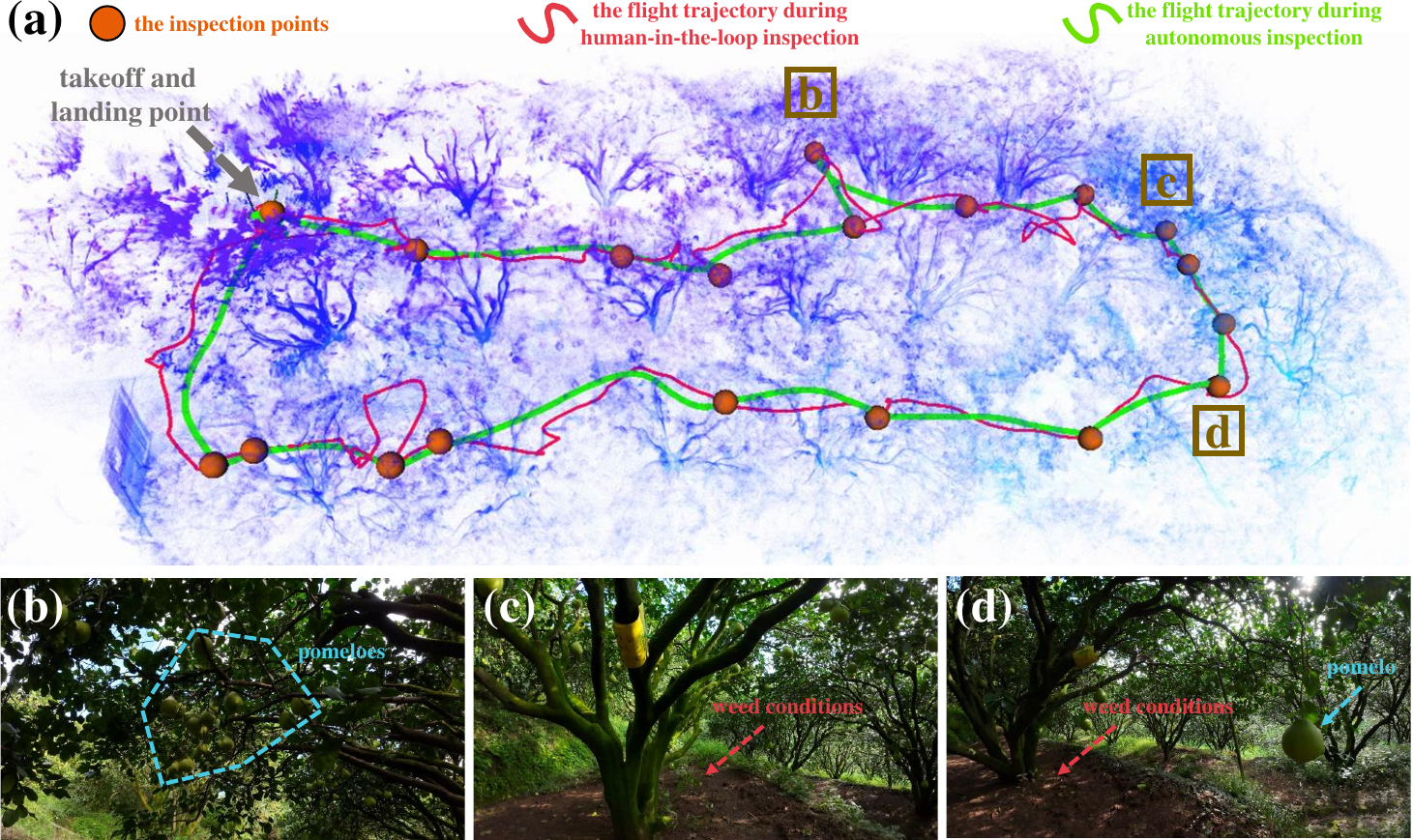}
    \caption{Overview of the quadrotor inspection system in agriculture task. (a) Point cloud map constructed during the inspection. (c) Image captured of the weed conditions beneath the pomelo trees. (d) Image captured of the weed conditions beneath the pomelo trees and the growth status of the pomelo.}
    \label{fig:agriculture}
\end{figure}

We performed a quadrotor inspection task for agricultural crops, specifically pomelo, in Meizhou City, Guangdong Province, China, known as the ``hometown of pomelo''\footnote{\href{https://en.wikipedia.org/wiki/Meizhou\#Food}{https://en.wikipedia.org/wiki/Meizhou\#Food}}. The orchard is situated in Changtian Township, Pingyuan County, characterized by hilly terrain, dense vegetation, and steep slopes. The primary objectives of this agricultural inspection task were to evaluate weed conditions beneath the pomelo trees (Fig. \ref{fig:agriculture}(b) and Fig. \ref{fig:agriculture}(d)) and assess the growth status of the pomelo (Fig. \ref{fig:agriculture}(c) and Fig. \ref{fig:agriculture}(d)).

\begin{table}[h]
    \centering
    \caption{Flight Data of Inspection for Agriculture Task}
    \label{tab:agriculture_data}
    \begin{tabular}{|c|c|c|c|c|}
        \hline
        Inspection Mode & Maximum & Average & Trajectory & Flight Time \\
         & Speed $(m/s)$ & Speed $(m/s)$ & Length $(m)$ & $(min : sec)$ \\
        \hline
        Human-in-the-loop & 2.26 & 0.54 & 128.60 & 3 : 58 \\
        \hline
        Autonomous & 2.20 & 0.82 & 92.52 & 1 : 53 \\
        \hline
    \end{tabular}
\end{table}

In this experiment, our quadrotor completed the full inspection workflow. The flight data recorded during the inspection is shown in Table \ref{tab:agriculture_data}. Compared to the human-in-the-loop inspection phase, the autonomous inspection phase achieved a 28\% reduction in trajectory length and a 53\% reduction in flight time. This improvement can be attributed to the dense environment within the orchard, where the UAV frequently rejected coarse commands from inexperienced operators during the human-in-the-loop inspection. In contrast, the autonomous inspection phase provided the UAV with greater freedom, enabling more precise and efficient navigation.

\begin{figure}[h]
    \centering
    \includegraphics[width=\textwidth]{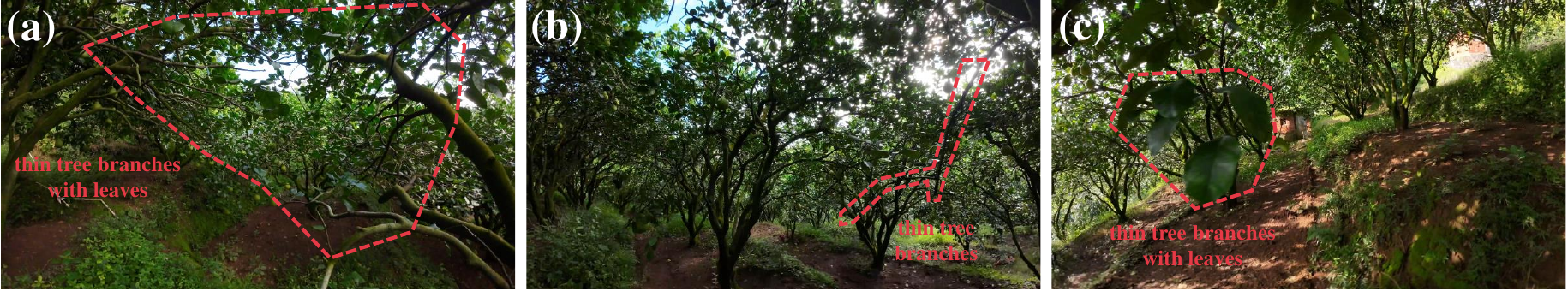}
    \caption{First-person view photos collected during autonomous inspection. Our quadrotor successfully avoided small obstacles, demonstrating the robustness of the navigation system.}
    \label{fig:agriculture_small}
\end{figure}

\begin{figure}[h]
    \centering
    \includegraphics[width=\textwidth]{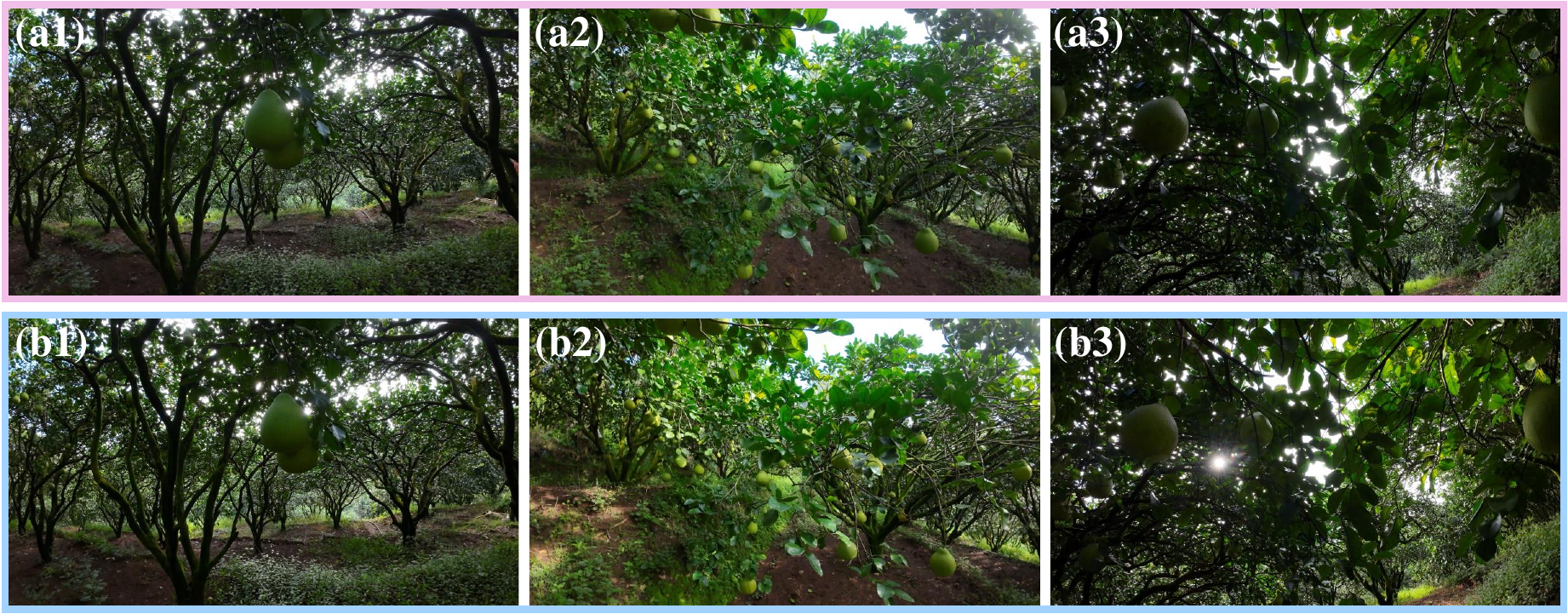}
    \caption{Comparison of photos collected at the same inspection points during human-in-the-loop inspection and autonomous inspection. (a) Images collected during human-in-the-loop inspection. (b) Images collected during autonomous inspection.}
    \label{fig:agriculture_bench}
\end{figure}

As shown in Fig. \ref{fig:agriculture}, during the autonomous inspection, the quadrotor autonomously navigated to the inspection points recorded during the human-in-the-loop phase, following the shortest path to collect data. Additionally, it safely avoided small objects (\eg, thin tree branches) in the environment, as demonstrated in Fig. \ref{fig:agriculture_small}. Furthermore, by leveraging the re-localization functionality of the localization module (Sec. \ref{sec:auto_lio}), the photos of the inspection targets collected during both the human-in-the-loop and autonomous phases were nearly identical, as shown in Fig. \ref{fig:agriculture_bench}.

Through efficient quadrotor inspection, we were able to accurately monitor the status of the agricultural crops, promptly identify potential issues, and provide critical data to support subsequent measures.

We invite the readers to view our third supplementary video\footnote{\href{https://youtu.be/pTsQbbMZJy4}{https://youtu.be/pTsQbbMZJy4}} for a clearer understanding of the entire process of the quadrotor executing the agriculture inspection task.


\subsection{Inspection for Factory Task}

\begin{figure}[h]
    \centering
    \includegraphics[width=\textwidth]{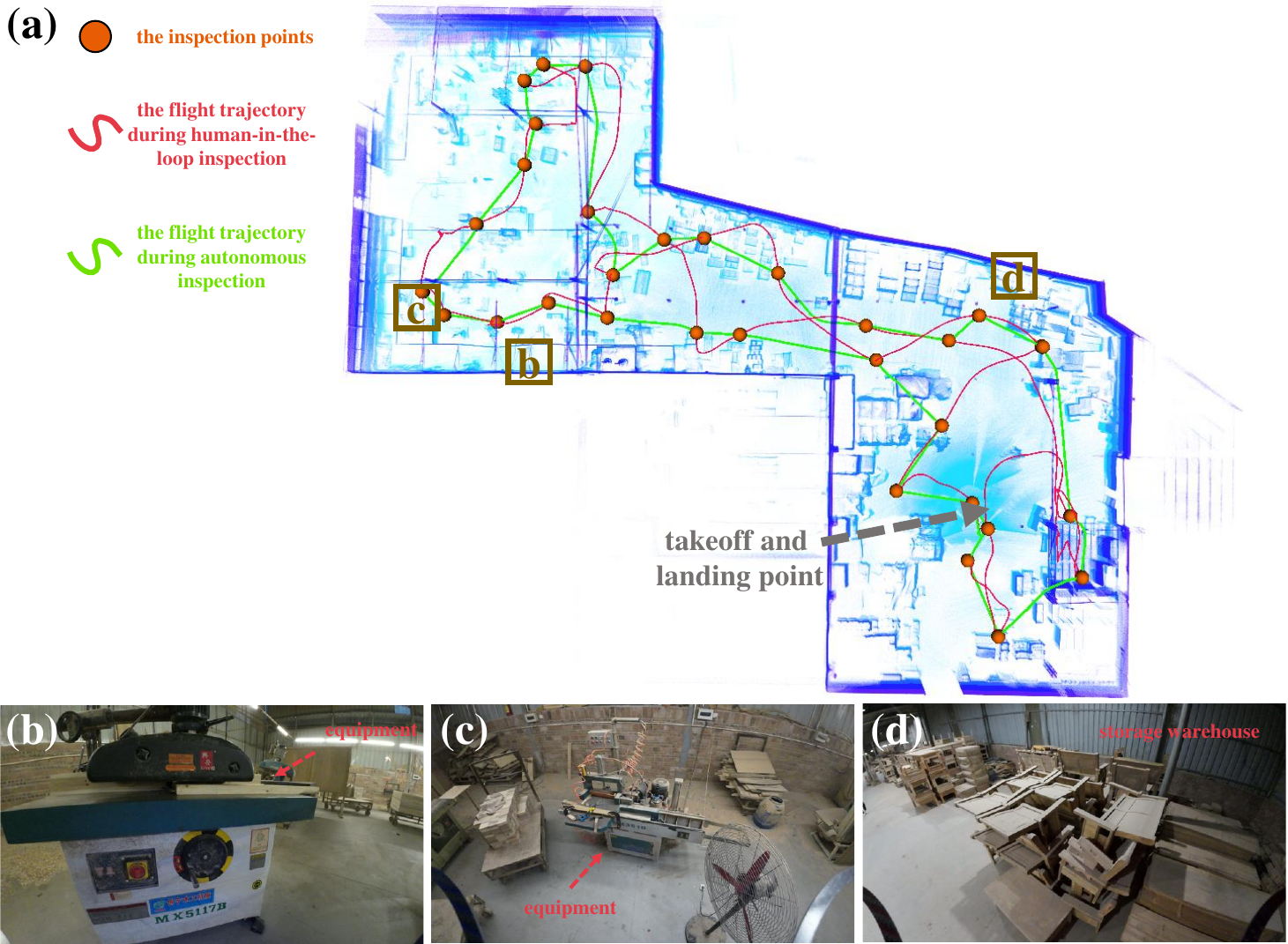}
    \caption{Overview of the quadrotor inspection system in factory task. (a) Point cloud map constructed during the inspection. (b-c) Image captured of the operational status of equipment in the processing area. (d) Image captured of the storage conditions of products in the storage area.}
    \label{fig:factory}
\end{figure}

Next, we conducted a quadrotor inspection task in a typical indoor factory environment. The factory, a standard furniture manufacturing facility, is divided into a storage area and a processing area. The primary inspection targets for this task were the operational status of equipment in the processing area (Fig. \ref{fig:factory}(b) and Fig. \ref{fig:factory}(c)) and the storage conditions of products in the storage area (Fig. \ref{fig:factory}(d)).

\begin{table}[h]
    \centering
    \caption{Flight Data of Inspection for Factory Task}
    \label{tab:factory_data}
    \begin{tabular}{|c|c|c|c|c|}
        \hline
        Inspection Mode & Maximum & Average & Trajectory & Flight Time \\
        & Speed $(m/s)$ & Speed $(m/s)$ & Length $(m)$ & $(min : sec)$ \\
        \hline
        Human-in-the-loop & 2.19 & 1.06 & 319.10 & 5 : 01 \\
        \hline
        Autonomous & 2.18 & 1.02 & 229.66 & 3 : 45 \\
        \hline
    \end{tabular}
\end{table}

In this experiment, our quadrotor completed the entire inspection workflow. The flight data recorded during the inspection is summarized in Table \ref{tab:factory_data}. Although the maximum flight speeds in both phases were similar, the efficiency of the autonomous inspection resulted in a 28\% reduction in trajectory length and a 25\% reduction in flight time compared to the human-in-the-loop inspection phase.

\begin{figure}[h]
    \centering
    \includegraphics[width=\textwidth]{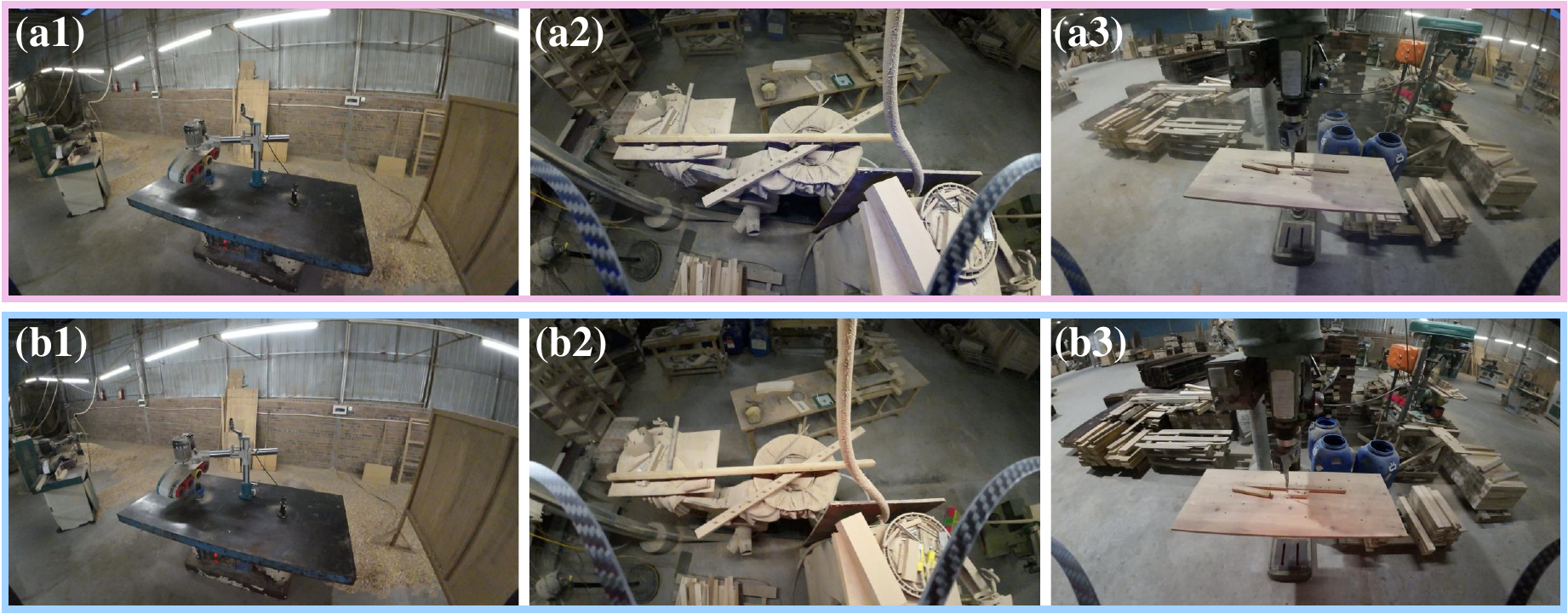}
    \caption{Comparison of photos collected at the same inspection points during human-in-the-loop inspection and autonomous inspection. (a) Images collected during human-in-the-loop inspection. (b) Images collected during autonomous inspection.}
    \label{fig:factory_bench}
\end{figure}

As illustrated in Fig. \ref{fig:factory}, during the autonomous inspection, the quadrotor autonomously navigated to the inspection points recorded during the human-in-the-loop phase, following the shortest path to collect data. Additionally, by leveraging the re-localization functionality of the localization module (Sec. \ref{sec:auto_lio}), the photos of the inspection targets collected during both the human-in-the-loop and autonomous phases were nearly identical, as shown in Fig. \ref{fig:factory_bench}.

Through efficient quadrotor inspection, we were able to accurately monitor the factory’s latest status, promptly identify potential safety hazards, and provide critical data to support subsequent production and maintenance activities.

We encourage readers to view our fourth supplementary video\footnote{\href{https://youtu.be/h1ToZjvdV8s}{https://youtu.be/h1ToZjvdV8s}} for a comprehensive demonstration of the quadrotor executing the factory inspection task.


\subsection{Inspection for Forestry Task}

\begin{figure}[h]
    \centering
    \includegraphics[width=\textwidth]{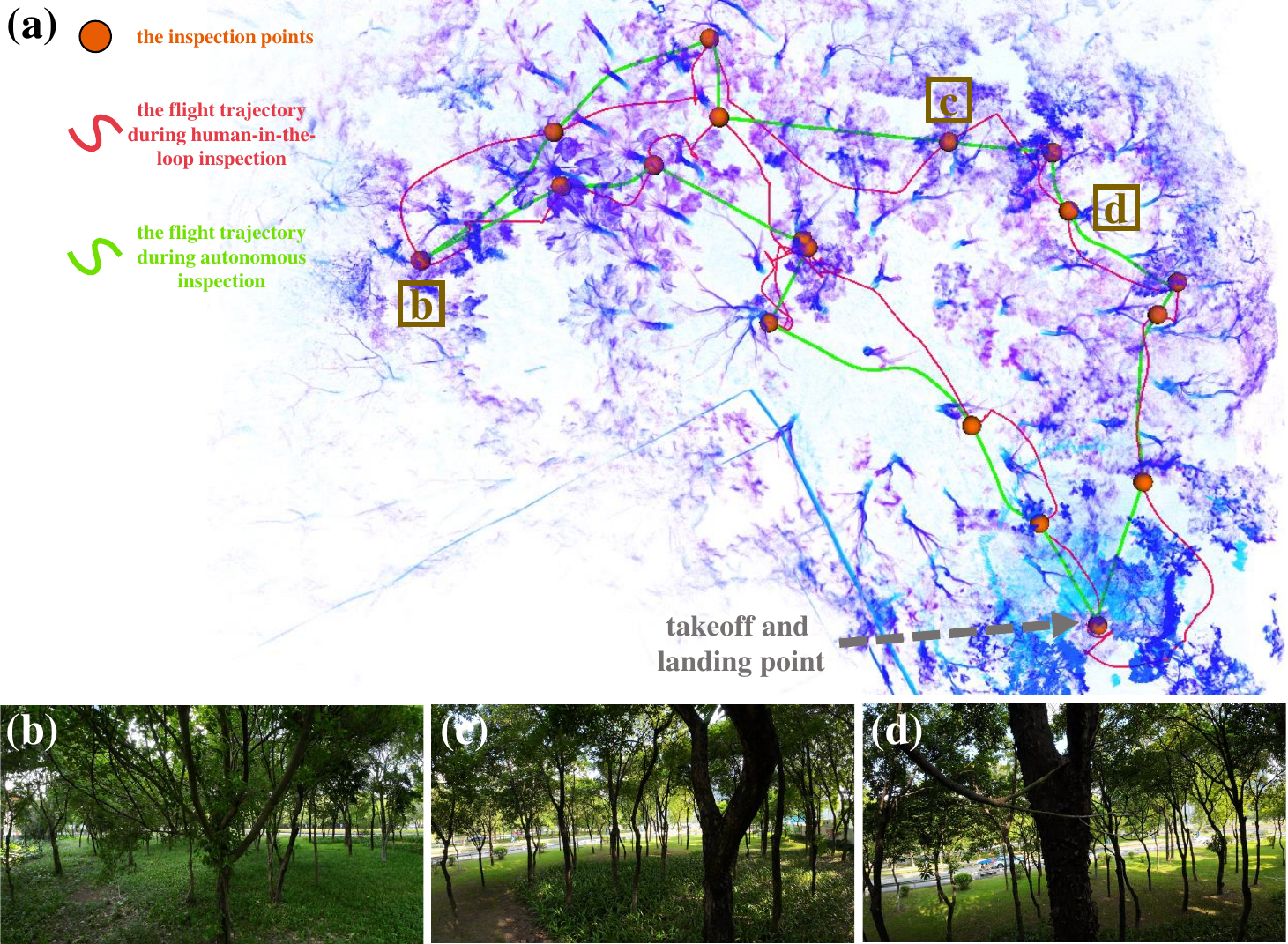}
    \caption{Overview of the quadrotor inspection system in forestry task. (a) Point cloud map constructed during the inspection. (b-d) Image captured of the growth status of the trees.}
    \label{fig:forestry}
\end{figure}

Finally, we conducted a quadrotor inspection task in a forest environment. The primary objective of this forestry inspection task was to assess the growth status of the trees.

\begin{table}[h]
    \centering
    \caption{Flight Data of Inspection for Forestry Task}
    \label{tab:forestry_data}
    \begin{tabular}{|c|c|c|c|c|}
        \hline
        Inspection Mode & Maximum & Average & Trajectory & Flight Time \\
         & Speed $(m/s)$ & Speed $(m/s)$ & Length $(m)$ & $(min : sec)$ \\
        \hline
        Human-in-the-loop & 2.25 & 0.84 & 257.74 & 5 : 06 \\
        \hline
        Autonomous & 2.20 & 1.23 & 173.88 & 2 : 21 \\
        \hline
    \end{tabular}
\end{table}

In this experiment, our quadrotor completed the entire inspection workflow. The flight data recorded during the inspection is presented in Table \ref{tab:forestry_data}. Although the maximum flight speeds in both phases were similar, the efficiency of the autonomous inspection resulted in a 33\% reduction in trajectory length and a 54\% reduction in flight time compared to the human-in-the-loop inspection phase.

\begin{figure}[h]
    \centering
    \includegraphics[width=\textwidth]{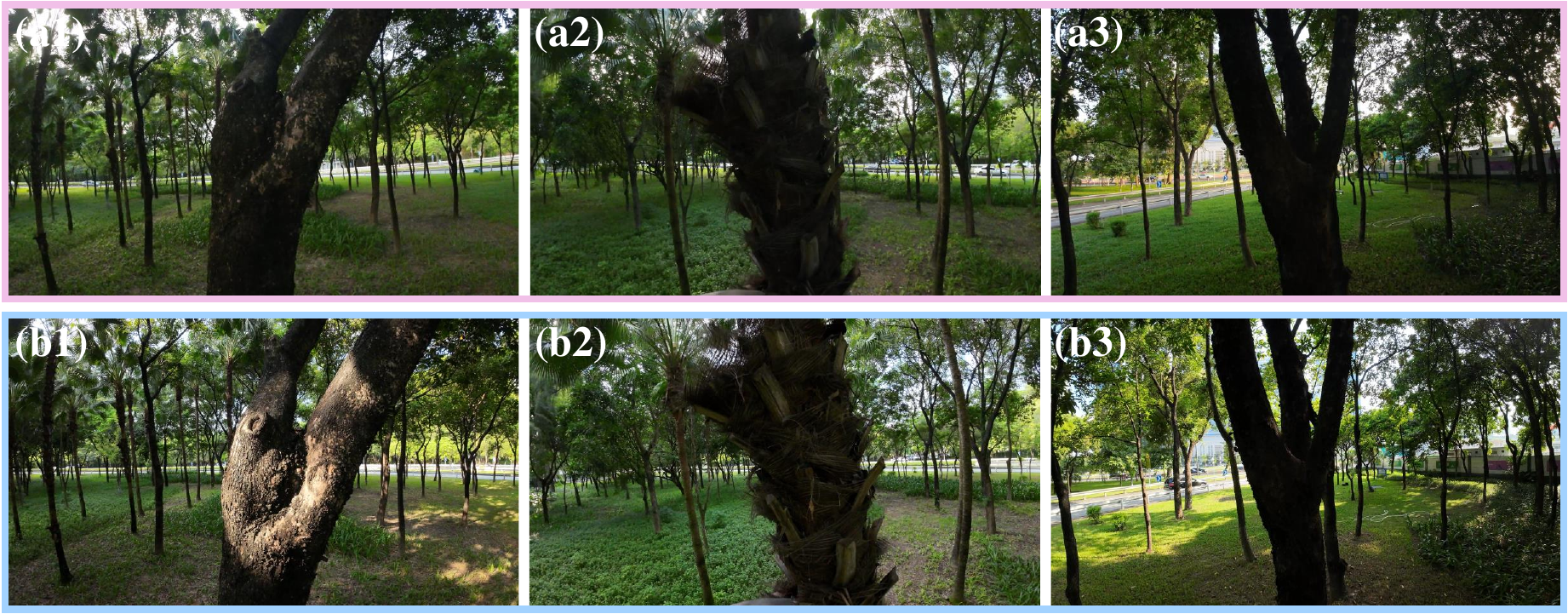}
    \caption{Comparison of photos collected at the same inspection points during human-in-the-loop inspection and autonomous inspection. (a) Images collected during human-in-the-loop inspection. (b) Images collected during autonomous inspection.}
    \label{fig:forestry_bench}
\end{figure}

As shown in Fig. \ref{fig:forestry}, during the autonomous inspection, the quadrotor autonomously navigated to the inspection points recorded during the human-in-the-loop phase, following the shortest path to collect data. Furthermore, by leveraging the re-localization functionality of the localization module (Sec. \ref{sec:auto_lio}), the photos of the inspection targets collected during the human-in-the-loop and autonomous phases were nearly identical, as illustrated in Fig. \ref{fig:forestry_bench}.

Through efficient quadrotor inspection, we were able to accurately monitor the latest status of the forested area, promptly identify potential safety hazards, and provide critical data to support subsequent forestry maintenance tasks.

We encourage readers to view our fifth supplementary video\footnote{\href{https://youtu.be/-Djl8-_2QCY}{https://youtu.be/-Djl8-\_2QCY}} for a detailed demonstration of the quadrotor performing the forestry inspection task.


\section{Conclusion}

This paper introduces a LiDAR-based quadrotor inspection system designed to perform various tasks in cluttered, unstructured, and GNSS-denied environments. The system operates in two phases: human-in-the-loop inspection and autonomous inspection, effectively integrating the capabilities of untrained pilots with an autonomous quadrotor to achieve efficient inspection task execution. The main features of this system include its capability to be operated by untrained pilots and its ability to autonomously conduct inspection tasks in diverse, unknown, GNSS-denied, and cluttered environments.

Extensive field experiments conducted in diverse environments, including slopes, landslides, factories, agricultural areas, and forests, validate the reliability and adaptability of the system. The results highlight significant reductions in trajectory length and flight time during autonomous inspections compared to human-in-the-loop operations, along with improved safety and reduced energy consumption. The system’s ability to autonomously avoid obstacles and collect high-quality inspection data further underscores its practicality for real-world applications.

By addressing limitations in existing UAV inspection systems, this work contributes to a comprehensive and scalable solution for UAV-based inspections, paving the way for advancements in industrial, environmental, and agricultural monitoring. Future efforts will aim to improve the scalability of the system and extend its applicability to more dynamic and complex scenarios, including rapidly changing or highly congested environments.

\subsubsection*{Acknowledgments}

The authors gratefully acknowledge DJI for fund support and Livox Technology for equipment support during the project. Special thanks are extended to GuiTian Xiao, Fushan Liu, and Guilu Xiao for their invaluable assistance with the experiments and for offering access to the experimental sites.


\end{document}